\documentclass[letterpaper,journal]{IEEEtran}
\usepackage{amsmath,amsfonts}
\usepackage{algorithm}
\usepackage{array}
\usepackage[caption=false,font=normalsize,labelfont=sf,textfont=sf]{subfig}
\usepackage{textcomp}
\usepackage{stfloats}
\usepackage{url}
\usepackage{verbatim}
\usepackage{graphicx}
\usepackage{cite}
\usepackage{booktabs,multirow}
\usepackage{amssymb}
\usepackage{algpseudocode}

\begin{document}

\title{S\textsuperscript{3}-Nav: Structured Spatial-Semantic Decision Making for Language-Guided Robot Navigation}

\author{Chaemin Lee\textsuperscript{1} and
Hae Min Cho\textsuperscript{1,*}%
\thanks{\textsuperscript{1}Chaemin Lee and Hae Min Cho are with the Department of AI, Gachon University, Seongnam 13120, Republic of Korea (e-mail: chaemin8600@gachon.ac.kr; hmcho9@gachon.ac.kr).}%
\thanks{\textsuperscript{*}Corresponding author: Hae Min Cho (e-mail: hmcho9@gachon.ac.kr).}%
}
\maketitle

\begingroup
\renewcommand{\thefootnote}{}
\footnotetext{%
\footnotesize
This work has been submitted to the IEEE for possible publication.
Copyright may be transferred without notice, after which this version may no longer be accessible.
}
\endgroup

\begin{abstract}
Language-guided robot navigation under partial observability requires robots to reason over spatial-semantic evidence accumulated across observations, particularly when goals involve spatial relations between objects.
This paper presents S\textsuperscript{3}-Nav, a structured spatial-semantic decision-making framework that integrates online mapping, persistent object--relation memory, and goal-conditioned navigation. 
Using RGB-D observations and SLAM, S\textsuperscript{3}-Nav maintains a persistent object--relation memory containing object records and confirmed relation records alongside an evolving navigation context containing exploration frontiers, without requiring a pre-existing map. 
A goal-conditioned evidence interface selects task-relevant evidence from this structured state and constructs a capacity-controlled decision view as part of the high-level policy input for exploration, target approach, and stopping decisions. 
The persistent memory remains available for subsequent reasoning and verification even when some of its records are excluded from the decision view. 
To ensure grounded execution, provisional policy decisions are independently validated against persistent object--relation memory and current geometric evidence before physical execution or task termination. 
Evaluations across multiple environments and object- and relation-goal tasks show improved closed-loop navigation reliability and reduced language-model token consumption with capacity-controlled evidence exposure under the evaluated inference configuration.
\end{abstract}


\section{Introduction}
\label{sec:introduction}

\begin{figure}[!t]
    \centering
    \includegraphics[width=\columnwidth]{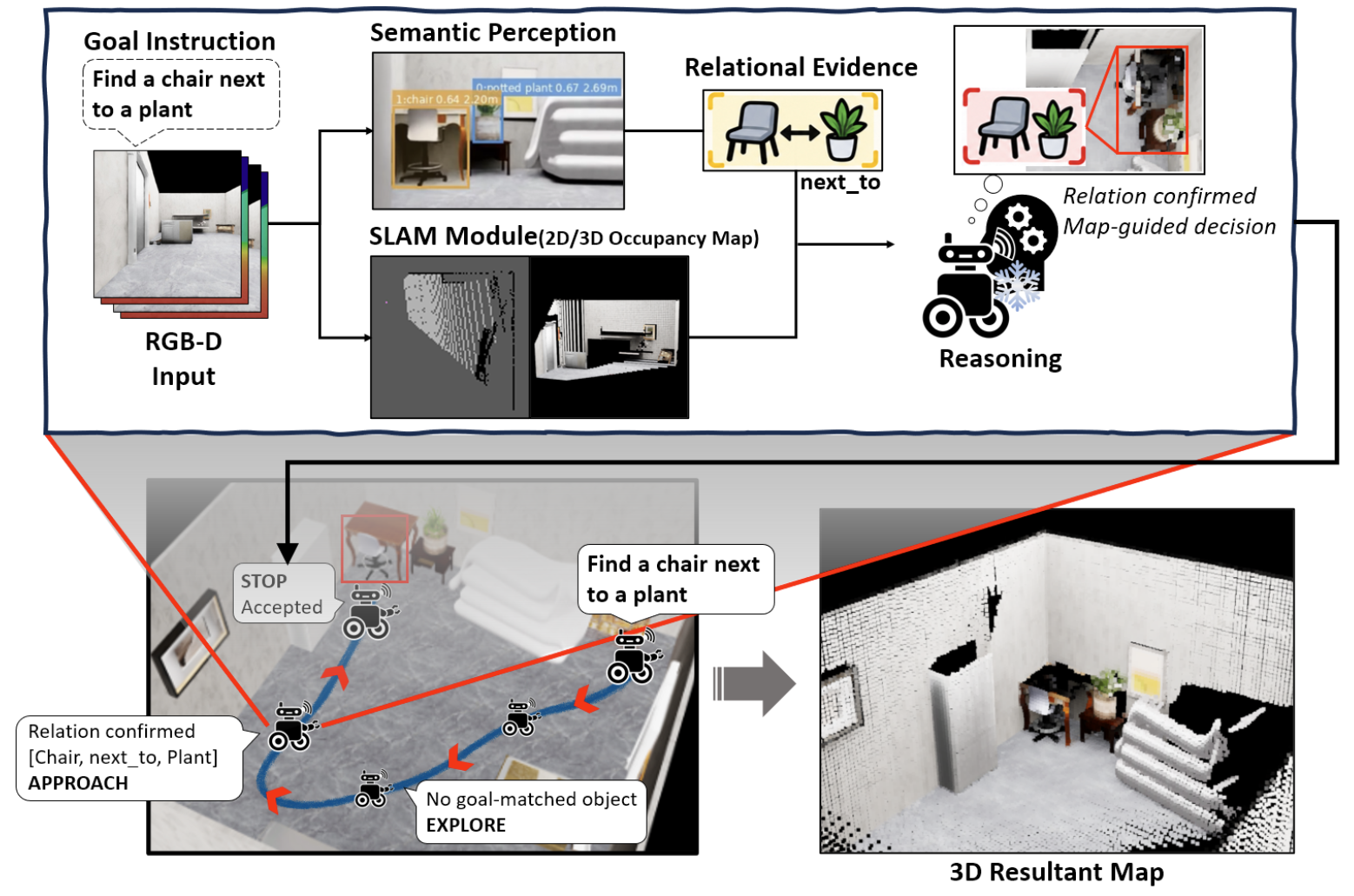}
    \caption{Qualitative example of relation-goal navigation for ``Find a chair next to a plant.''
    The sequence illustrates frontier-guided exploration, target-directed approach using accumulated spatial-semantic evidence, and termination following semantic and geometric validation.}
    \label{fig:qualitative_example}
\end{figure}

\IEEEPARstart{L}{anguage-guided} robot navigation often involves making decisions based on partial observations and spatial-semantic evidence accumulated over previous navigation steps \cite{zhou2024navgpt,yin2024sg}.
This is particularly apparent in relational instructions such as \textit{``find a chair next to a plant.''}
Detecting a chair provides a possible navigation target, but does not establish whether that chair satisfies the requested relation.
The plant may be observed from a different viewpoint, and even after a matching pair is identified, the robot still needs geometric evidence that it has reached an appropriate stopping position.
Fig.~\ref{fig:qualitative_example} illustrates the transition from exploration to target approach and terminal validation.

Prior research has explored language-conditioned reasoning, persistent spatial-semantic representations, and hierarchical navigation control as complementary capabilities for interpreting goals, retaining observations, and executing actions \cite{driess2023palm,gu2024conceptgraphs,cheng2024navila}.
In online navigation without a pre-existing map, these capabilities need to operate as the environment representation evolves.
Persistent object records retain map-frame positions across observations, allowing spatial predicates to be evaluated for object pairs observed at different times \cite{wani2020multion,gu2024conceptgraphs,honerkamp2024language,yin2024sg}.
Such evidence may remain useful after the corresponding objects leave the current camera view.

This setting presents two distinct requirements for language-guided navigation.
The first is to preserve spatial-semantic evidence accumulated across observations while selecting goal-relevant information for immediate policy decisions.
The second is to distinguish evidence that supports exploration or target approach from evidence sufficient to authorize task completion.

To address these requirements, we present S\textsuperscript{3}-Nav, a framework that uses RGB-D observations and SLAM to maintain a structured decision state comprising a goal schema, persistent object--relation memory, and current navigation context.
This structure organizes accumulated spatial-semantic evidence into explicit object and relation records while separating persistent memory from the evolving navigation context.
A goal-conditioned evidence interface selects and ranks task-relevant records to construct a capacity-controlled decision view without pruning persistent memory.
The high-level policy uses this view as part of its input to propose navigation intents, while an independent execution-validation mechanism applies distinct evidence requirements to target approach and task completion.

The main contributions of this work are:
\begin{itemize}
\item A goal-conditioned evidence interface that ranks object, relation, and frontier records to construct a capacity-controlled decision view while preserving persistent object--relation memory for subsequent grounding and validation.

\item An execution-validation mechanism that applies goal-dependent evidence requirements to target approach and task completion.
Relational stopping requires confirmed goal-matching relations and recent target-associated range evidence, independently of the decision-view capacity.

\item An experimental evaluation across HSSD and InternScenes examining the effects of high-level policy selection, decision-view capacity, persistent memory, and frontier records on closed-loop navigation performance and reasoning cost.  

\end{itemize}

\section{Related Work}
\label{sec:related_work}

\subsection{Spatial-Semantic Mapping and Memory}
\label{subsec:related_mapping}

Persistent spatial-semantic representations enable robots to retain environmental evidence beyond the current observation.
SemanticFusion~\cite{mccormac2017semanticfusion} integrates semantic predictions into persistent 3D maps, whereas Hydra~\cite{hughes2022hydra} organizes accumulated spatial information into hierarchical scene graphs.
ConceptGraphs~\cite{gu2024conceptgraphs} further supports object-centric representations of open-vocabulary objects and their spatial relationships.
For language-grounded navigation, VLMaps~\cite{huang2023visual} associates visual-language features with 3D maps, while HOV-SG~\cite{werby2024hierarchical} employs floor--room--object hierarchies to organize navigable environments.
Task relevance is also considered by Clio~\cite{maggio2024clio}, which constructs compact 3D scene graphs by selecting semantic information relevant to downstream tasks.
These approaches demonstrate different ways of accumulating, structuring, and selecting spatial-semantic information.
S\textsuperscript{3}-Nav focuses on separating persistent object--relation memory from the capacity-controlled decision view constructed for immediate policy decisions.
This separation limits policy-visible evidence while preserving accumulated object and relation records for subsequent grounding and validation.

\subsection{Object-Goal and Language-Guided Navigation}
\label{subsec:related_navigation}

Object-goal and language-guided navigation combine semantic exploration with language-conditioned target selection.
Goal-oriented semantic exploration~\cite{chaplot2020object} uses object-level semantic cues to guide target search, while LM-Nav~\cite{shah2023lmnav} grounds language instructions through visual landmarks and pretrained navigation models.
VLFM~\cite{yokoyama2024vlfm} combines frontier exploration with language-grounded value maps to guide zero-shot object navigation.
SG-Nav~\cite{yin2024sg} organizes accumulated scene information into hierarchical graphs, selects subgraphs for LLM-based frontier evaluation, and employs graph-based re-perception to improve goal detection reliability.
These approaches support navigation through semantic target selection and grounding, but relational instructions require distinguishing a plausible target from one that satisfies the requested spatial predicate.
S\textsuperscript{3}-Nav makes this distinction by allowing object records and transient relation cues to guide exploration or target approach while requiring confirmed goal-matching relation evidence for relation-goal satisfaction.

\subsection{Grounded Language-Model Reasoning}
\label{subsec:related_grounded_reasoning}

Grounded language-model control commonly separates high-level semantic reasoning from feasibility checking and physical execution.
Systems such as SayCan~\cite{ichter2023saycan} and Inner Monologue~\cite{huang2023inner} illustrate this separation by combining language-based reasoning with independent mechanisms for feasibility assessment or interaction with the environment.
SayPlan~\cite{rana2023sayplan} further retrieves task-relevant subgraphs for language-model planning and uses iterative replanning to address infeasible actions.
A similar hierarchy appears in language-guided navigation, where pretrained models select high-level actions and lower-level modules execute them \cite{shah2023lmnav,zhou2024navgpt,cheng2024navila}.
However, separating planning from execution does not by itself specify when task completion should be accepted.
S\textsuperscript{3}-Nav makes this boundary explicit by treating high-level policy outputs as provisional navigation intents that must be independently validated by the execution layer.
The execution layer resolves grounded references and determines whether task termination can be accepted.

\section{Method}
\label{sec:method}

S\textsuperscript{3}-Nav organizes its decision process by accumulating spatial-semantic evidence across observations, selecting goal-relevant information for policy decisions, and independently validating proposed navigation intents.
This separation preserves accumulated evidence for subsequent reasoning while distinguishing evidence that guides navigation from evidence sufficient to authorize task completion.
In this section, we first present the overall decision architecture and describe how online mapping and structured-state construction support the accumulation of spatial-semantic evidence.
We then explain how goal-conditioned decision-view construction supports high-level policy decisions and how execution validation governs grounded actions and task termination.

\subsection{Structured Spatial-Semantic Decision Architecture}
\label{sec:method_overview}

\begin{figure*}[!t]
\centering
\includegraphics[width=\textwidth]{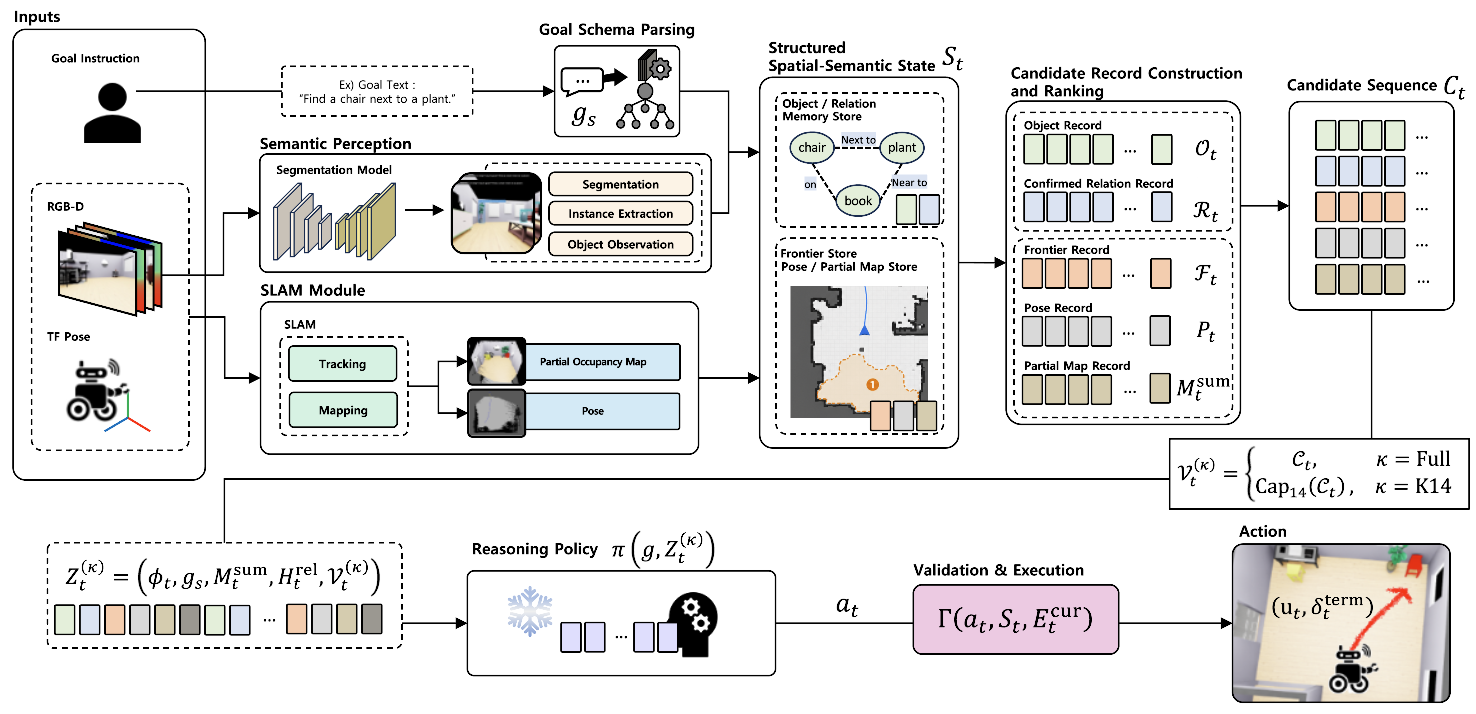}
\caption{Overview of the S\textsuperscript{3}-Nav decision architecture.
Online perception and mapping update the structured decision state $S_t$, the goal-conditioned evidence interface $\Psi_{\kappa}$ constructs a policy input containing a capacity-controlled decision view, and the execution layer $\Gamma$ validates provisional navigation intents using persistent object--relation memory and current geometric evidence.} 
\label{fig:system_pipeline}
\end{figure*}

The S\textsuperscript{3}-Nav architecture connects online perception, goal-conditioned policy decisions, and execution validation through a structured decision state.
At each decision step, the system processes RGB-D observations to update the robot pose and partial occupancy map and to obtain map-grounded semantic observations.
The structured decision state $S_t$ combines persistent object--relation memory $\mathcal{B}_t$ with the current navigation context, including exploration frontiers. 
The goal-conditioned evidence interface $\Psi_{\kappa}$ constructs the policy input $Z_t^{(\kappa)}$, including a capacity-controlled decision view $\mathcal{V}_t^{(\kappa)}$ derived from the ordered candidate sequence $C_t$.
Here, $\kappa$ specifies the record-capacity setting.
The high-level policy $\pi$ uses this input to propose a navigation intent $a_t$ with a structured-record reference when needed.

The proposed intent is evaluated by an execution function $\Gamma$ using the structured state $S_t$ and current geometric evidence $E_t^{\mathrm{cur}}$.
S\textsuperscript{3}-Nav separates controlling the evidence exposed to the high-level policy from validating the resulting provisional navigation intents.
The former controls which records are exposed to the high-level policy without pruning persistent memory, while the latter validates provisional navigation intents against persistent object--relation memory and current geometric evidence independently of the decision-view capacity.
The closed-loop information flow in Fig.~\ref{fig:system_pipeline} is formalized as
\begin{equation}
\begin{aligned}
Z_t^{(\kappa)}
&= \Psi_{\kappa}\!\left(S_t,\phi_t,H_t^{\mathrm{rel}}\right),\\
a_t
&= \pi\!\left(g,Z_t^{(\kappa)}\right),\\
(\mathbf{u}_t,\delta_t^{\mathrm{term}})
&= \Gamma\!\left(a_t,S_t,E_t^{\mathrm{cur}}\right).
\end{aligned}
\label{eq:closed_loop_architecture}
\end{equation}
Here, $g$ is the language instruction, $\phi_t$ is the navigation phase, and $H_t^{\mathrm{rel}}$ summarizes progress toward a relation-goal.
The outputs $\mathbf{u}_t$ and $\delta_t^{\mathrm{term}}$ denote the executable motion command and an accepted terminal event, respectively, with $\delta_t^{\mathrm{term}}=\varnothing$ when no termination is accepted.
Algorithm~\ref{alg:s3nav_main} details the state update, policy invocation, and validation sequence.

\begin{algorithm}[t]
\caption{S\textsuperscript{3}-Nav Closed-Loop Decision Procedure}
\label{alg:s3nav_main}
\footnotesize
\begin{algorithmic}[1]
\Require Instruction \(g\), high-level policy \(\pi\), record capacity \(\kappa\)
\State \(g_s \gets \Call{ParseGoalSchema}{g}\)
\State Initialize \(S_0\)
\For{each decision step \(t\)}
\State Acquire RGB-D observations and update perception, localization, and occupancy mapping
\State Update frontier candidates, persistent object--relation memory, and map summary  
\State $\mathcal{B}_t \gets (\mathcal{O}_t,\mathcal{R}_t)$
\State $\mathcal{N}_t \gets (\mathcal{F}_t,P_t,M_t^{\mathrm{sum}})$
\State $S_t \gets (g_s,\mathcal{B}_t,\mathcal{N}_t)$
\State \(\mathcal{C}_t \gets \Call{ConstructCandidateSequence}{S_t}\)
\State \(\mathcal{V}_t^{(\kappa)} \gets \Call{ConstructDecisionView}{\mathcal{C}_t,\kappa}\)
\State Update navigation phase \(\phi_t\) and relation-search summary \(H_t^{\mathrm{rel}}\)
\State \(Z_t^{(\kappa)} \gets (\phi_t,g_s,M_t^{\mathrm{sum}},H_t^{\mathrm{rel}},\mathcal{V}_t^{(\kappa)})\)
\State \(a_t \gets \pi(g,Z_t^{(\kappa)})\)
\State Obtain current execution evidence \(E_t^{\mathrm{cur}}\)
\State \((\mathbf{u}_t,\delta_t^{\mathrm{term}}) \gets \Gamma(a_t,S_t,E_t^{\mathrm{cur}})\)
\If{\(\delta_t^{\mathrm{term}}\neq\varnothing\)}
\State \textbf{break}
\EndIf
\State Execute \(\mathbf{u}_t\)
\EndFor
\end{algorithmic}
\end{algorithm}

\subsection{Online Localization, Mapping, and Frontier Construction}
\label{sec:rgbd_slam_frontier}

S\textsuperscript{3}-Nav maintains geometric context from online RGB-D observations without requiring a pre-existing environment map.
Localization and occupancy mapping provide the robot pose $P_t$ and an evolving partial map $\mathcal{M}_t$, establishing a common spatial reference for object grounding and navigation.
The boundaries between observed free space and unexplored regions yield frontier candidates $\mathcal{F}_t$, which represent potential exploration destinations rather than semantic targets for task completion \cite{yamauchi1997frontier}.
The map summary $M_t^{\mathrm{sum}}$ contains the explored-space ratio, frontier count, and accumulated path length, while the full occupancy map remains available to the navigation stack without being included in the policy input.

\subsection{Structured Spatial-Semantic Decision State}
\label{sec:spatial_state}
\label{sec:goal_schema_parsing}
\label{sec:rgbd_semantic_perception}
\label{sec:spatial_memory_repository}

The state separates spatial-semantic evidence retained across observations from navigation quantities updated at the current step:
\begin{equation}
\begin{aligned}
S_t &= (g_s,\mathcal{B}_t,\mathcal{N}_t),\\
\mathcal{B}_t &= (\mathcal{O}_t,\mathcal{R}_t),\\
\mathcal{N}_t &= (\mathcal{F}_t,P_t,M_t^{\mathrm{sum}}).
\end{aligned}
\label{eq:state}
\end{equation}
The goal schema $g_s=(o_t,p,o_r,o_s,c_r)$ specifies the target category, requested predicate, reference category, approach-anchor category, and whether relational verification is required.
The persistent object--relation memory $\mathcal{B}_t$ preserves object records and confirmed relation records across observations, whereas the navigation context $\mathcal{N}_t$ reflects the robot's evolving pose, exploration frontiers, and map progress.
This separation allows previously acquired semantic evidence to remain available as the local navigation context changes.

As in object-centric spatial-semantic mapping \cite{gu2024conceptgraphs}, each object record in $\mathcal{O}_t$ maintains a persistent identity, semantic category, map-frame position, confidence, and observation history.
Detected instances are localized by back-projecting valid aligned RGB-D mask pixels and transforming their coordinate-wise median into the active map frame.
Association with an existing object requires category agreement and spatial proximity in that frame, after which the stored identity is retained and its position and observation metadata are updated.
A new record is created when no admissible match exists, while observations without reliable depth support are not admitted as grounded object anchors.
These persistent identities let an object remain an approach candidate after it leaves the current camera view.

Consistent with incremental 3D scene-graph representations \cite{wu2021scenegraphfusion}, each confirmed relation record in $\mathcal{R}_t$ links persistent subject and reference object identities through a spatial predicate that has passed geometric verification.
Along with the two object references, it retains the physical navigation anchor, supporting geometry, confidence, and the history of geometric verification.
For \texttt{next\_to}, geometric verification evaluates the horizontal separation between localized object pairs and their vertical separation when height estimates are available.
The admissible separation depends on whether the object pair matches the target and reference categories specified by the relation-goal instruction.
A broader tolerance is applied to goal-matching pairs to reduce the rejection of potentially relevant relations under approximate object localization, while stricter criteria for other pairs limit incidental relations in persistent memory.
For \texttt{on}, the test additionally considers vertical ordering and projected support or a proximity-based fallback.
Object co-occurrence alone cannot create a confirmed relation, and updating robot-relative anchor geometry does not renew its verification timestamp.
An unrefreshed relation eventually expires, while the constituent object records remain available for subsequent reasoning.
This distinction is important because the presence of a remembered target and reference supplies possible navigation anchors, whereas a confirmed relation supplies evidence that the requested predicate has been geometrically verified.

For example, the instruction \textit{``find a chair next to a plant''} produces $g_s=(\texttt{chair},\texttt{next\_to},\texttt{plant},\texttt{chair},1)$.
A chair record $q_c$ and plant record $q_p$ can guide search or approach without establishing that the goal relation is satisfied.
Once the pair passes geometric verification, the memory admits a confirmed relation $q_r=(q_c,\texttt{next\_to},q_p)$ that retains the chair as the approach and stopping anchor. 
The goal-conditioned evidence interface can therefore expose object records, transient relation cues representing unconfirmed spatial relationships, or confirmed relation records for different decisions, while the persistent memory maintains the underlying grounded references.
The semantic requirements for accepting a terminal decision are defined separately in Sec.~\ref{sec:execution_layer}.

\subsection{Decision View Construction and High-Level Policy}
\label{sec:structured_token_construction}
\label{sec:reasoning_policy}

At each decision step, the goal-conditioned evidence interface constructs an ordered candidate sequence $\mathcal{C}_t$ from the structured state $S_t$. 
Frontier candidates from $\mathcal{F}_t$ are represented as frontier records when included in $\mathcal{C}_t$.
For relation-goals, the interface also generates transient relation cues from localized, goal-relevant object pairs.
Unlike confirmed relations in $\mathcal{R}_t$, these candidates indicate possible relationships without establishing their validity.
They retain references to their source objects and can be regenerated as the structured decision state changes. 
When a transient relation cue is selected for approach, its source-object references are used to resolve the physical navigation anchor without promoting the cue to a confirmed relation in $\mathcal{R}_t$.
This allows the policy to consider incomplete relational evidence during exploration and target approach without treating it as sufficient evidence for task completion.

Consistent with task-driven scene abstraction and subgraph selection \cite{maggio2024clio,rana2023sayplan}, candidate selection follows a staged priority structure that prioritizes goal-schema records, goal- and relation-relevant object and relation evidence, frontier records, and remaining context, in that order.
The implementation admits up to two goal-schema records, extends the selected prefix to at most ten with goal-related records, and then considers up to five frontier records before filling remaining positions. 
Within each stage, a deterministic candidate-ranking heuristic considers goal compatibility, relational relevance, record type, semantic salience, observation confidence, and robot-relative distance.
Goal-compatible object records and confirmed relation records receive greater priority than proximity alone, since the nearest observed object need not satisfy the requested instruction.
The ranking coefficients and stage limits are fixed heuristic parameters chosen to express these priorities rather than learned scores or theoretically optimized values.
For a given structured state $S_t$, all capacity settings share the same ranked candidate sequence $\mathcal{C}_t$.

The decision view is defined as
\begin{equation}
\mathcal{V}_t^{(\kappa)}
=
\begin{cases}
\mathcal{C}_t,
& \kappa=\mathrm{Full},\\
\operatorname{Cap}_{\kappa}(\mathcal{C}_t),  
& \kappa\in\mathbb{N}_{>0},
\end{cases}
\label{eq:record_view}
\end{equation}
where $\kappa\in\mathbb{N}_{>0}\cup\{\mathrm{Full}\}$ specifies the decision-view capacity setting, and $\operatorname{Cap}_{\kappa}$ retains at most the first $\kappa$ records of the ordered candidate sequence when $\kappa$ is finite.
For a fixed candidate sequence, any finite-capacity decision view is an order-preserving prefix of the Full view, restricting policy-visible records without discarding the underlying evidence from persistent memory $\mathcal{B}_t$.

The navigation phase $\phi_t$ is updated from goal-compatible evidence and frontier availability, while the relation-search summary $H_t^{\mathrm{rel}}$ tracks confirmed relation matches and previously inspected unconfirmed targets.
Together, they help the policy distinguish exploration, target approach, and provisional stopping candidates without independently authorizing termination.
The complete policy input is
\begin{equation}
Z_t^{(\kappa)}
=
\left(
\phi_t,\,
g_s,\,
M_t^{\mathrm{sum}},\,
H_t^{\mathrm{rel}},\,
\mathcal{V}_t^{(\kappa)}
\right).
\label{eq:decision_context}
\end{equation}
The capacity constraint applies only to $\mathcal{V}_t^{(\kappa)}$, leaving the remaining policy-input components available regardless of the selected record count.

\begin{figure*}[t]
\centering
\includegraphics[width=1.00\textwidth]{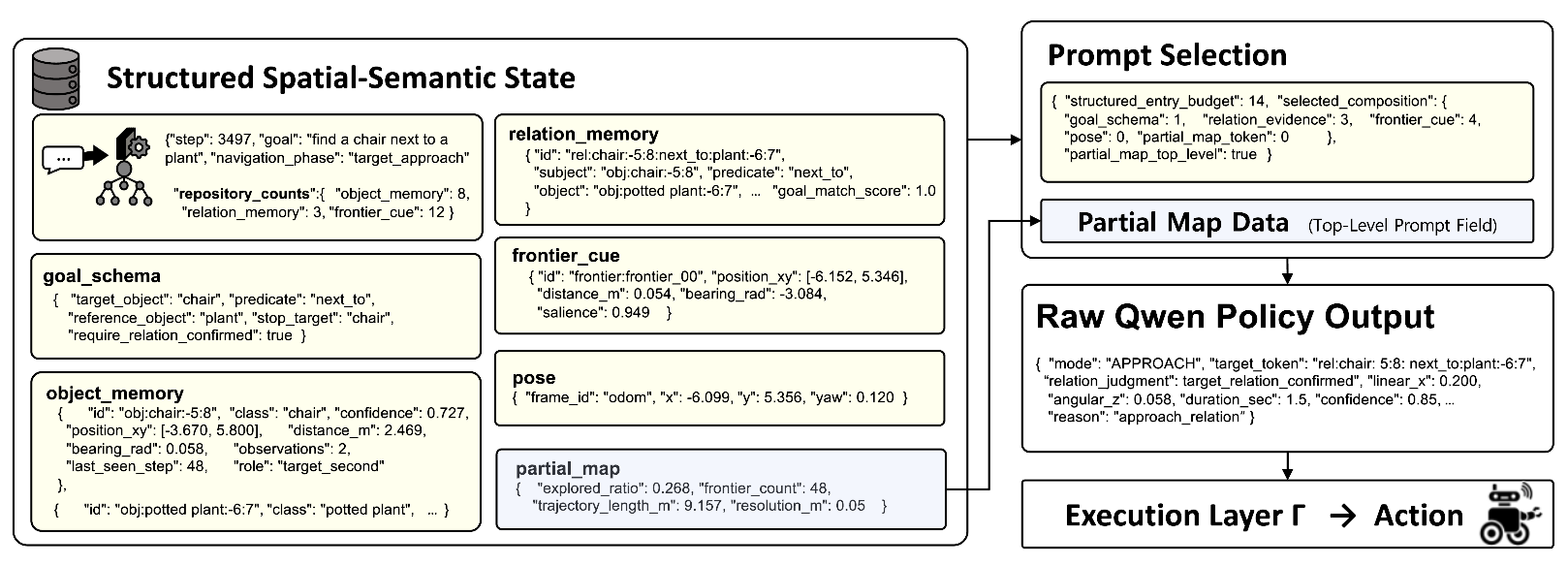}
\caption{Goal-conditioned evidence selection and high-level policy interface.
Ranked object, relation, and frontier records form a capacity-controlled decision view, while structured-record references connect provisional policy outputs to execution validation.}
\label{fig:structured_interface}
\end{figure*}

Given the instruction $g$ and policy input $Z_t^{(\kappa)}$, the high-level policy proposes a navigation intent with a structured-record reference when applicable: 
\begin{equation}
\begin{aligned}
a_t
&=
\pi\!\left(g,Z_t^{(\kappa)}\right)\\
&=
\left(
m_t,q_t,v_t,\omega_t,\eta_t
\right),
\end{aligned}
\label{eq:reasoning_policy_action}
\end{equation}
where $m_t\in\{\textsc{Explore},\textsc{Approach},\textsc{Hold},\textsc{Stop}\}$ denotes the navigation mode, $q_t$ is the selected structured-record reference, $v_t$ and $\omega_t$ are proposed linear and angular velocities, respectively, subject to predefined bounds, and $\eta_t$ contains auxiliary policy information.
Qwen generates the provisional intent from the serialized instruction and policy input, whereas the heuristic applies fixed priority rules to the selected records, with both outputs subject to the same execution validation. 
Fig.~\ref{fig:structured_interface} illustrates how the selected references connect policy decisions to grounded execution.

\subsection{Decision Validation and Execution Authority}
\label{sec:execution_layer}
\label{sec:termination_logging}
The high-level policy produces a provisional mode and structured-record reference rather than an immediately authorized action.
A remembered target position can guide approach even after the target leaves the current view, but cannot establish whether the robot currently satisfies the required stopping range.
Termination therefore requires recent target-associated range evidence and, for relation-goals, a confirmed goal-matching relation supporting the selected physical target.
For \textsc{Explore}, the execution layer resolves a frontier destination, whereas \textsc{Approach} resolves the physical navigation anchor from a selected object record, transient relation cue, or confirmed relation record.
A \textsc{Hold} proposal does not initiate target-directed motion.
In each case, the execution function $\Gamma$ resolves the proposed reference against $S_t$ and applies current geometric and safety constraints from $E_t^{\mathrm{cur}}$.

Let $y_q$ denote the semantic category of the physical navigation anchor associated with stopping reference $q$.
The stopping reference $q$ may itself be an object record, transient relation cue, or confirmed relation record.
The predicate $\operatorname{RelValid}(q,g_s,\mathcal{R}_t)$ equals one when the persistent relation memory $\mathcal{R}_t$ contains a confirmed goal-matching relation supporting the physical target resolved from stopping reference $q$, and zero otherwise.
The semantic admissibility of a stopping reference is
\begin{equation}
\begin{aligned}
\operatorname{StopAdmissible}(q,g_s,\mathcal{R}_t)
&= \mathbb{I}[y_q=o_s]\Bigl(1-c_r\\
&\qquad+c_r\,\operatorname{RelValid}(q,g_s,\mathcal{R}_t)\Bigr).
\end{aligned}
\label{eq:stop_admissible}
\end{equation}
This test establishes semantic eligibility but does not independently authorize termination.

The terminal-evaluation range is obtained from recent target-associated depth observations as
\begin{equation}
d_t^{\mathrm{eval}}
= \Psi_{\mathrm{term}}\!\left(q_t,S_t,I_t^{\mathrm{d}}\right),
\label{eq:terminal_range}
\end{equation}
where $I_t^{\mathrm{d}}$ is the current depth observation and $\Psi_{\mathrm{term}}$ resolves an eligible target-associated range.
A measurement is eligible only if its associated target observation was refreshed no earlier than the immediately preceding memory update.
The resolver prioritizes target-associated surface measurements computed from valid target-mask depth points using the tenth percentile of their horizontal distances in the robot base frame.

For approach, a valid terminal-evaluation range is used when available, with remembered map-grounded geometry serving as a fallback.
\begin{equation}
d_t^{\mathrm{app}}=
\begin{cases}
d_t^{\mathrm{eval}}, & \text{if valid},\\
\Psi_{\mathrm{map}}(q_t,S_t), & \text{otherwise},
\end{cases}
\label{eq:approach_range}
\end{equation}
where $\Psi_{\mathrm{map}}$ estimates the approach range from the remembered map-frame position of the physical navigation anchor and the current robot pose.

Combining semantic admissibility with recent target-associated range evidence gives the terminal gate
\begin{equation}
\begin{aligned}
\chi_t^{\mathrm{stop}}=\mathbb{I}\Bigl[
&\operatorname{StopAdmissible}(q_t,g_s,\mathcal{R}_t)=1\\
&\land\; d_t^{\mathrm{eval}}\ \text{is valid}\\
&\land\; d_{\min}\leq d_t^{\mathrm{eval}}\leq d_{\max}
\Bigr],
\end{aligned}
\label{eq:primary_stop_validity}
\end{equation}
with $d_{\min}=0.5\,\mathrm{m}$ and $d_{\max}=0.9\,\mathrm{m}$ in the evaluated tasks.
A policy-generated \textsc{Stop} is accepted only if $\chi_t^{\mathrm{stop}}=1$.
The same condition can also authorize termination during deterministic standoff control after an \textsc{Approach} proposal.
Such execution-triggered termination is distinguished from a policy-requested stop in the logs.

Motion safety is handled separately from semantic and terminal authorization.
A depth-based guard evaluates obstacle proximity and depth validity before applying the proposed velocity, and may hold, reduce speed, steer, or trigger bounded recovery.
These safety overrides do not promote tentative object or relation evidence into an admissible stopping condition.

\section{Experiments}
\label{sec:experiments}

\subsection{Experimental Setup}
\label{sec:experimental_setup}

The primary experiments are conducted in indoor household environments from the HSSD dataset \cite{khanna2024habitat} using Isaac Sim.
An additional evaluation uses three InternScenes navigation tasks \cite{zhong2026internscenes} to examine whether the observed policy and record-capacity patterns are consistent across environments.
Both evaluations use a differential-drive Ranger Mini equipped with an RGB-D camera and share the same semantic perception, mapping, persistent object--relation memory, and deterministic execution procedures unless otherwise stated.
Semantic perception uses YOLO26n-seg~\cite{jocher2026ultralytics}, while RTAB-Map~\cite{labbe2019rtabmap} provides the map-frame pose and partial occupancy map under Estimated localization.
All runs are executed on a workstation with two NVIDIA RTX 5080 GPUs.
For policy and record-capacity comparisons, evidence construction is held fixed, including a $0.75\,\mathrm{m}$ XY object-association threshold, an 80-update expiration limit for unrefreshed relations, and the same predicate-specific verification and deterministic ranking rules.
For \texttt{next\_to}, the horizontal and vertical separation thresholds are $(1.80,0.85)\,\mathrm{m}$ for goal-matching pairs and $(0.75,0.55)\,\mathrm{m}$ for other pairs, with the vertical criterion applied only when height estimates are available.

The high-level language-model policy uses Qwen3-4B-Thinking-2507 \cite{yang2025qwen3} served through a local vLLM-compatible endpoint.
Each request is subject to a 6,000-token limit shared by the prompt and completion.
When the constructed prompt exceeds 9,000 characters, the inference interface applies a concise prompt that shortens instruction text and selected record attributes.
Because Full exposes the complete candidate sequence whereas K14 retains at most 14 records, the two views can differ in whether this adaptation is triggered.
Requests may also encounter context-length overflow when the adapted input exceeds the model's token budget.
Consequently, the comparison evaluates record capacity under the implemented inference interface rather than isolating the effect of record count alone.
K14 was chosen as the default finite capacity before the subsequent sensitivity study and is not assumed to be optimal.

The primary comparison evaluates Heuristic Full, Heuristic K14, Qwen Full, and Qwen K14 under Estimated localization.
For the localization comparison, Qwen K14 instead receives simulator GT Odometry as the pose source while retaining the same mapping backend and downstream navigation stack.
The ground-truth robot pose is treated as an odom-frame pose under simulator world--odom alignment and transformed into the active map frame using the current map-to-odom estimate.

\subsection{Tasks and Evaluation Protocol}
\label{sec:tasks_protocol}

The task set includes object-goal and relation-goal instructions with different requirements for exploration, object identification, and relational verification.
Object-goals require navigation to an object of the specified target category, while relation-goals additionally require the correct target--predicate--reference combination.
For relation-goals, the target object serves as the physical approach and stopping anchor.
The five HSSD scenarios and three InternScenes tasks are evaluated over five repeated runs per primary condition.
Initial poses, instructions, and designated instances are matched across corresponding runs, while random seeds are matched when the corresponding procedures support seed control.
Continuous measurements are aggregated from run-level results and reported as means and sample standard deviations.
Table~\ref{tab:evaluation_scenarios} describes the tasks and their initial visibility conditions.

\begin{table}[t]
\centering
\caption{Object- and relation-goal navigation scenarios in HSSD and InternScenes.}
\label{tab:evaluation_scenarios}
\footnotesize
\setlength{\tabcolsep}{2.5pt}
\renewcommand{\arraystretch}{1.10}
\begin{tabular}{@{}lp{0.50\columnwidth}cc@{}}
\toprule
Task & Instruction & Initial & Distractor \\
\midrule
\multicolumn{4}{@{}l}{\textit{HSSD}} \\
S1 & find a chair next to a plant & Seen & No \\
S2 & find a chair next to a plant & Unseen & No \\
S3 & find a vase on a chair & Seen & No \\
S4 & find a plant & Unseen & No \\
S5 & find a chair next to a plant & Unseen & Yes \\
\midrule
\multicolumn{4}{@{}l}{\textit{InternScenes}} \\
I1 & find a chair & Seen & No \\
I2 & find a plant next to a chair & Seen & No \\
I3 & find a chair next to a plant & Unseen & No \\
\bottomrule
\end{tabular}
\end{table}

Initial visibility is determined by whether the designated instance is within the initial camera field of view and sensing range with an unobstructed line of sight, independently of detection success.
For object-goals, \emph{Seen} indicates that the target is initially observable.
For relation-goals, both the target and reference objects must be initially observable to qualify as \emph{Seen}.
The paired HSSD relation-goal tasks with different initial visibility distinguish navigation with immediately accessible evidence from navigation requiring exploration and memory across observations.
The unseen distractor case further tests whether persistent object--relation memory retains competing object and relation evidence and whether the goal-conditioned evidence interface selects task-relevant candidates. 
The \texttt{on} relation-goal task and the object-goal task probe different semantic requirements, while InternScenes provides an additional environment with both initially seen and unseen tasks.

\subsection{Evaluation Metrics}
\label{sec:evaluation_metrics}

The primary evaluation concerns whether an episode terminates at the correct semantic target, satisfies the requested spatial relation when applicable, and reaches the required stopping range.
Semantic Standoff Success Rate (SSR) measures success according to the navigation system's recorded semantic, relational, and range assessments, extending established navigation success criteria based on target identification and stopping proximity \cite{chaplot2020object, anderson2018evaluation}.
Ground-Truth Semantic Standoff Success Rate (GT-SSR) applies the corresponding requirements using simulator-ground-truth target identity, relation, and geometry, extending established evaluation criteria based on target identity and stopping proximity \cite{krantz2023navigating}.
We treat GT-SSR as the principal task-level metric because agreement with the controller's own stopping evidence does not necessarily imply ground-truth correctness.
Reporting SSR alongside GT-SSR makes that distinction observable without treating online terminal acceptance as an independent success guarantee.

For episode $e$, let $T_e^{\mathrm{valid}}$ indicate valid ground-truth terminal geometry, $M_e^{\mathrm{GT}}$ indicate correct ground-truth semantic and, when applicable, relational agreement, and $d_{e,\mathrm{surf}}^{\mathrm{GT}}$ denote ground-truth surface range to the physical target anchor.
With target standoff $d^{\star}=0.7\,\mathrm{m}$ and tolerance $\tau_{\mathrm{standoff}}=0.2\,\mathrm{m}$, GT-SSR is computed as
\begin{equation}
\begin{aligned}
\mathrm{GT\text{-}SSR}
&=
\frac{1}{N}\sum_{e=1}^{N}
\mathbb{I}\Bigl[
T_e^{\mathrm{valid}}
\land M_e^{\mathrm{GT}}\\
&\qquad\land
\left|d_{e,\mathrm{surf}}^{\mathrm{GT}}-d^{\star}\right|
\leq\tau_{\mathrm{standoff}}
\Bigr].
\end{aligned}
\label{eq:gt_ssr}
\end{equation}
Both SSR and GT-SSR use the complete episode set, including unsuccessful attempts and timeouts.
Episodes without valid ground-truth terminal geometry are counted as failures in GT-SSR.
The auxiliary count $N_{\mathrm{GT\text{-}valid}}$ indicates how many episodes have valid ground-truth terminal geometry for interpretation of conditional measures.

Relation Success Rate (RSR) isolates predicate satisfaction at termination for relation-goal episodes with a valid relation verdict, providing a separate assessment of spatial predicates used in relation-goal navigation \cite{boularias2016learning}.
Because its denominator excludes episodes without an eligible verdict, a high RSR need not indicate successful navigation or correct standoff.
Standoff Navigation Error (SNE) and GT-SNE measure absolute deviation from $d^{\star}$ using available online and ground-truth terminal ranges, respectively.
Unlike GT-SSR, these error measures are conditional on valid measurements and are therefore interpreted alongside success rates and the corresponding validity counts.
Trajectory Length (TL) is computed from the simulator-ground-truth trajectory, and Episode Duration includes time until termination or timeout.
Navigation-policy call counts and token consumption quantify language-model reasoning demand, while Policy Latency and end-to-end (E2E) Decision Latency measure policy processing and the full observe-to-command interval, respectively.
Translational and rotational Relative Pose Error (RPE) assess pose accuracy in the secondary localization comparison.

At eligible target observations, the sensor-based surface range is computed as the tenth percentile of horizontal distances to valid target-mask depth points expressed in the robot base frame.
The ground-truth range is the minimum horizontal distance from the robot base to the target's world-frame axis-aligned bounding box.
Surface-range MAE and RMSE characterize the discrepancy between these measurements, with $N_{\mathrm{surface}}$ counting valid observation steps.
Unlike GT-SNE, these diagnostics concern intermediate target observations collected along condition-specific trajectories.

\begin{table*}[!t]
\centering
\caption{Navigation performance of Qwen and heuristic policies under Full and K14 decision views with Estimated localization.
Each task--condition pair includes five episodes, with SSR and GT-SSR computed over all episodes.
RSR uses valid relation verdicts, while range errors are computed from valid terminal measurements.
Bold indicates the best value for each metric within a task, excluding $N_{\mathrm{GT\text{-}valid}}$, with ties included.}
\label{tab:qwen_vs_heuristic}
\footnotesize
\setlength{\tabcolsep}{1.15pt}
\renewcommand{\arraystretch}{1.06}
\begin{tabular}{@{}llcccccccc@{}}
\toprule
Task & Policy / view & SSR $\uparrow$ & GT-SSR $\uparrow$ & $N_{\mathrm{GT\text{-}valid}}$ & RSR $\uparrow$ & SNE (m) $\downarrow$ & GT-SNE (m) $\downarrow$ & TL (m) $\downarrow$ & Duration (s) $\downarrow$ \\
\midrule
\multicolumn{10}{@{}l}{\textit{HSSD}} \\
\midrule
\multirow{4}{*}{S1} & Heuristic Full & $\mathbf{1.000}$ & $\mathbf{1.000}$ & $5$ & $\mathbf{1.000}$ & $0.172\pm0.017$ & $0.148\pm0.018$ & $3.27\pm0.02$ & $25.30\pm1.15$ \\
& Heuristic K14 & $\mathbf{1.000}$ & $\mathbf{1.000}$ & $5$ & $\mathbf{1.000}$ & $\boldsymbol{0.170\pm0.023}$ & $\boldsymbol{0.145\pm0.023}$ & $3.28\pm0.03$ & $25.40\pm1.85$ \\
& Qwen Full & $\mathbf{1.000}$ & $\mathbf{1.000}$ & $5$ & $\mathbf{1.000}$ & $0.173\pm0.009$ & $0.148\pm0.009$ & $3.27\pm0.01$ & $24.60\pm0.82$ \\
& Qwen K14 & $\mathbf{1.000}$ & $\mathbf{1.000}$ & $5$ & $\mathbf{1.000}$ & $0.186\pm0.010$ & $0.161\pm0.010$ & $\boldsymbol{3.26\pm0.01}$ & $\boldsymbol{23.70\pm0.76}$ \\
\midrule
\multirow{4}{*}{S2} & Heuristic Full & $\mathbf{1.000}$ & $\mathbf{1.000}$ & $5$ & $\mathbf{1.000}$ & $0.149\pm0.027$ & $0.135\pm0.027$ & $\boldsymbol{2.54\pm0.32}$ & $17.10\pm3.07$ \\
& Heuristic K14 & $\mathbf{1.000}$ & $\mathbf{1.000}$ & $5$ & $\mathbf{1.000}$ & $\boldsymbol{0.141\pm0.032}$ & $\boldsymbol{0.120\pm0.031}$ & $3.19\pm0.08$ & $21.10\pm1.14$ \\
& Qwen Full & $0.800$ & $0.600$ & $3$ & $\mathbf{1.000}$ & $0.169\pm0.022$ & $0.163\pm0.030$ & $2.69\pm0.56$ & $136.30\pm259.24$ \\
& Qwen K14 & $\mathbf{1.000}$ & $\mathbf{1.000}$ & $5$ & $\mathbf{1.000}$ & $0.155\pm0.022$ & $0.141\pm0.022$ & $2.60\pm0.39$ & $\boldsymbol{16.70\pm3.17}$ \\
\midrule
\multirow{4}{*}{S3} & Heuristic Full & $0.400$ & $0.400$ & $5$ & $\mathbf{1.000}$ & $0.186\pm0.004$ & $0.142\pm0.004$ & $4.17\pm0.25$ & $49.20\pm14.83$ \\
& Heuristic K14 & $\mathbf{0.800}$ & $\mathbf{0.800}$ & $5$ & $\mathbf{1.000}$ & $0.101\pm0.063$ & $\boldsymbol{0.057\pm0.062}$ & $4.40\pm0.29$ & $\boldsymbol{39.00\pm11.80}$ \\
& Qwen Full & $\mathbf{0.800}$ & $\mathbf{0.800}$ & $5$ & $\mathbf{1.000}$ & $0.101\pm0.036$ & $0.058\pm0.037$ & $4.88\pm0.73$ & $43.70\pm10.76$ \\
& Qwen K14 & $\mathbf{0.800}$ & $\mathbf{0.800}$ & $4$ & $0.800$ & $\boldsymbol{0.084\pm0.011}$ & $0.061\pm0.049$ & $\boldsymbol{3.61\pm1.91}$ & $41.10\pm10.63$ \\
\midrule
\multirow{4}{*}{S4} & Heuristic Full & $0.800$ & $0.800$ & $5$ & -- & $0.156\pm0.006$ & $\boldsymbol{0.009\pm0.008}$ & $7.81\pm2.54$ & $69.10\pm32.68$ \\
& Heuristic K14 & $0.800$ & $0.800$ & $5$ & -- & $0.178\pm0.012$ & $0.034\pm0.012$ & $8.12\pm3.34$ & $69.20\pm31.84$ \\
& Qwen Full & $0.400$ & $0.400$ & $5$ & -- & $\boldsymbol{0.107\pm0.029}$ & $0.035\pm0.031$ & $8.30\pm2.77$ & $94.50\pm34.97$ \\
& Qwen K14 & $\mathbf{1.000}$ & $\mathbf{1.000}$ & $5$ & -- & $0.144\pm0.026$ & $0.015\pm0.013$ & $\boldsymbol{6.33\pm0.63}$ & $\boldsymbol{52.80\pm5.96}$ \\
\midrule
\multirow{4}{*}{S5} & Heuristic Full & $0.200$ & $0.200$ & $1$ & $0.200$ & $0.131\pm0.000$ & $\mathbf{0.034}$ & $\boldsymbol{8.38\pm4.83}$ & $493.10\pm239.04$ \\
& Heuristic K14 & $0.600$ & $0.400$ & $3$ & $0.750$ & $0.151\pm0.063$ & $0.086\pm0.012$ & $10.04\pm1.68$ & $280.40\pm291.84$ \\
& Qwen Full & $0.400$ & $0.200$ & $1$ & $0.600$ & $\boldsymbol{0.033\pm0.012}$ & $0.120$ & $8.72\pm2.29$ & $380.90\pm300.32$ \\
& Qwen K14 & $\mathbf{1.000}$ & $\mathbf{1.000}$ & $5$ & $\mathbf{1.000}$ & $0.160\pm0.017$ & $0.064\pm0.018$ & $10.69\pm0.05$ & $\boldsymbol{75.40\pm1.88}$ \\
\midrule
\multicolumn{10}{@{}l}{\textit{InternScenes}} \\
\midrule
\multirow{4}{*}{I1} & Heuristic Full & $\mathbf{1.000}$ & $0.800$ & $5$ & -- & $\boldsymbol{0.1524\pm0.0366}$ & $0.0372\pm0.0210$ & $3.7589\pm0.1641$ & $\boldsymbol{33.20\pm0.76}$ \\
& Heuristic K14 & $\mathbf{1.000}$ & $\mathbf{1.000}$ & $5$ & -- & $0.1632\pm0.0328$ & $0.0376\pm0.0079$ & $3.7003\pm0.0588$ & $33.40\pm1.08$ \\
& Qwen Full & $\mathbf{1.000}$ & $\mathbf{1.000}$ & $5$ & -- & $0.1724\pm0.0075$ & $\boldsymbol{0.0332\pm0.0106}$ & $3.8562\pm0.3213$ & $38.50\pm4.78$ \\
& Qwen K14 & $\mathbf{1.000}$ & $\mathbf{1.000}$ & $5$ & -- & $0.1801\pm0.0141$ & $0.0485\pm0.0161$ & $\boldsymbol{3.6541\pm0.0194}$ & $34.00\pm0.79$ \\
\midrule
\multirow{4}{*}{I2} & Heuristic Full & $\mathbf{1.000}$ & $\mathbf{1.000}$ & $5$ & $\mathbf{1.000}$ & $0.1980\pm0.0014$ & $0.1498\pm0.0023$ & $4.0274\pm0.0352$ & $\boldsymbol{32.70\pm0.91}$ \\
& Heuristic K14 & $\mathbf{1.000}$ & $\mathbf{1.000}$ & $5$ & $\mathbf{1.000}$ & $0.1986\pm0.0012$ & $0.1508\pm0.0009$ & $4.0329\pm0.0536$ & $33.50\pm2.03$ \\
& Qwen Full & $\mathbf{1.000}$ & $\mathbf{1.000}$ & $5$ & $\mathbf{1.000}$ & $\boldsymbol{0.1966\pm0.0020}$ & $\boldsymbol{0.1491\pm0.0024}$ & $\boldsymbol{4.0182\pm0.0529}$ & $33.90\pm1.08$ \\
& Qwen K14 & $\mathbf{1.000}$ & $\mathbf{1.000}$ & $5$ & $\mathbf{1.000}$ & $0.1985\pm0.0010$ & $0.1508\pm0.0017$ & $4.1272\pm0.1111$ & $37.00\pm2.29$ \\
\midrule
\multirow{4}{*}{I3} & Heuristic Full & $0.600$ & $0.600$ & $3$ & $0.600$ & $\boldsymbol{0.1561\pm0.0155}$ & $0.0245\pm0.0174$ & $17.4776\pm6.5961$ & $309.28\pm267.22$ \\
& Heuristic K14 & $0.800$ & $0.600$ & $3$ & $0.800$ & $0.1717\pm0.0071$ & $0.0384\pm0.0074$ & $13.3745\pm5.6158$ & $204.24\pm222.90$ \\
& Qwen Full & $0.400$ & $0.400$ & $2$ & $\mathbf{1.000}$ & $0.1733\pm0.0366$ & $\boldsymbol{0.0228\pm0.0278}$ & $12.0163\pm3.1444$ & $415.33\pm253.34$ \\
& Qwen K14 & $\mathbf{1.000}$ & $\mathbf{1.000}$ & $5$ & $\mathbf{1.000}$ & $0.1667\pm0.0439$ & $0.0478\pm0.0140$ & $\boldsymbol{11.6510\pm2.4471}$ & $\boldsymbol{100.12\pm30.11}$ \\
\bottomrule
\end{tabular}
\end{table*}

\begin{table*}[!t]
\centering
\caption{Secondary localization and target-surface range diagnostics under Qwen K14.
Each task--condition pair includes five attempts.
RPE is reported as mean $\pm$ sample standard deviation, while surface-range errors are aggregated over valid target-associated observations from condition-specific trajectories.
$N_{\mathrm{surface}}$ denotes the total number of valid target-associated observations.}
\label{tab:localization_comparison}
\footnotesize
\setlength{\tabcolsep}{3pt}
\renewcommand{\arraystretch}{1.12}
\begin{tabular}{@{}llcccccc@{}}
\toprule
Suite & Pose source
& \shortstack{GT-SSR\\$\uparrow$}
& \shortstack{Transl. RPE\\(m) $\downarrow$}
& \shortstack{Rot. RPE\\(rad) $\downarrow$}
& \shortstack{Surface MAE\\(m) $\downarrow$}
& \shortstack{Surface RMSE\\(m) $\downarrow$}
& $N_{\mathrm{surface}}$ \\
\midrule
\multirow{2}{*}{HSSD}
& Estimated
& $\mathbf{0.960}$
& $0.1527\pm0.1638$
& $0.0726\pm0.0828$
& $\mathbf{0.0695}$
& $\mathbf{0.0878}$
& $537$ \\
& GT Odometry
& $0.840$
& $\boldsymbol{0.0034\pm0.0011}$
& $\boldsymbol{0.0030\pm0.0011}$
& $0.0752$
& $0.0967$
& $568$ \\
\midrule
\multirow{2}{*}{InternScenes}
& Estimated
& $\mathbf{1.000}$
& $0.1178\pm0.1365$
& $0.0582\pm0.0568$
& $\mathbf{0.1022}$
& $\mathbf{0.1076}$
& $567$ \\
& GT Odometry
& $\mathbf{1.000}$
& $\boldsymbol{0.0251\pm0.0561}$
& $\boldsymbol{0.0121\pm0.0320}$
& $0.1082$
& $0.1140$
& $525$ \\
\bottomrule
\end{tabular}
\end{table*}

\begin{table*}[t]
\centering
\caption{Effects of persistent memory and explicit frontier records on HSSD S5 using Qwen K14 under Estimated localization. 
Each condition includes five attempts, with SSR and GT-SSR computed over all attempts and RSR over valid relation verdicts.
TL and duration use available run records, excluding one incomplete memory-ablation attempt.
Bold indicates the highest GT-SSR.}
\label{tab:component_ablation}
\footnotesize
\setlength{\tabcolsep}{4.5pt}
\renewcommand{\arraystretch}{1.10}
\begin{tabular}{@{}lcccccc@{}}
\toprule
Condition & SSR $\uparrow$ & GT-SSR $\uparrow$ & $N_{\mathrm{GT\text{-}valid}}$ & RSR $\uparrow$ & TL (m) $\downarrow$ & Duration (s) $\downarrow$ \\
\midrule
Complete (Qwen K14) & $1.000$ & $\mathbf{1.000}$ & $5$ & $1.000$ & $10.688\pm0.047$ & $75.40\pm1.88$ \\
w/o Persistent Memory & $0.000$ & $0.000$ & $0$ & $0.250$ & $19.263\pm6.615$ & $600.00\pm0.00$ \\
w/o Explicit Frontier Records & $0.800$ & $0.800$ & $4$ & $1.000$ & $8.790\pm2.794$ & $177.70\pm236.09$ \\ 
\bottomrule
\end{tabular}
\end{table*}

\subsection{Comparison with Heuristic Policy}
\label{sec:qwen_vs_heuristic}

Table~\ref{tab:qwen_vs_heuristic} shows that Qwen K14 achieves higher GT-SSR than Qwen Full across all initially unseen tasks, with the largest improvements on HSSD S4 and S5, as well as InternScenes I3, while GT-SSR remains unchanged on initially seen tasks.
The heuristic exhibits a different pattern, with smaller or absent improvements on unseen tasks but a noticeable gain on the initially seen \texttt{on} task.
These differences suggest that the effect of decision-view capacity depends on both task conditions and the high-level policy.
A possible explanation lies in the evidence demands of tasks requiring exploration or discrimination among accumulated candidates.
In the initially unseen relation-goal conditions, the robot may encounter the target and reference at different times, and a policy decision must account for remembered objects, possible relational matches, and remaining exploration destinations.
When the relevant objects are available from the initial view and the policy input is relatively simple, Full and K14 can lead to similar outcomes.
The observed results are consistent with the usefulness of goal-relevant evidence selection, although the experiment does not isolate individual record selections as the cause of particular navigation actions.

Qwen generates a navigation mode and structured-record reference from the serialized instruction and policy input, whereas the heuristic selects the same action fields through fixed rules over the selected evidence. 
This action-selection heuristic is distinct from the candidate-ranking heuristic, which is shared by both policies.
Both policies share the same evidence construction and execution validation, allowing their decision rules to be compared under the evaluated decision-view capacities.
The heuristic is an internal alternative decision rule rather than a previously published navigation baseline.
For InternScenes I3, Qwen Full attains a perfect RSR among eligible relation verdicts but substantially lower episode-level GT-SSR.
This discrepancy illustrates why conditional relation and distance metrics must be interpreted alongside GT-SSR.

\subsection{Record-Capacity Sensitivity}
\label{sec:context_localization_analysis}

The Full--K14 comparison cannot establish whether a smaller view is always preferable, since a finite prefix may omit useful candidates as well as irrelevant records.
We therefore evaluate finite decision-view capacities $\kappa\in\{3,7,14\}$ and Full on HSSD S5 and InternScenes I3, where navigation requires exploration and relational decisions over evidence accumulated at different times. 
All capacities use the same retained memory, staged ranking, and execution validation, with each finite view taking an order-preserving prefix of the common candidate sequence.

\begin{figure}[!t]
\centering
\includegraphics[width=1.0\linewidth]{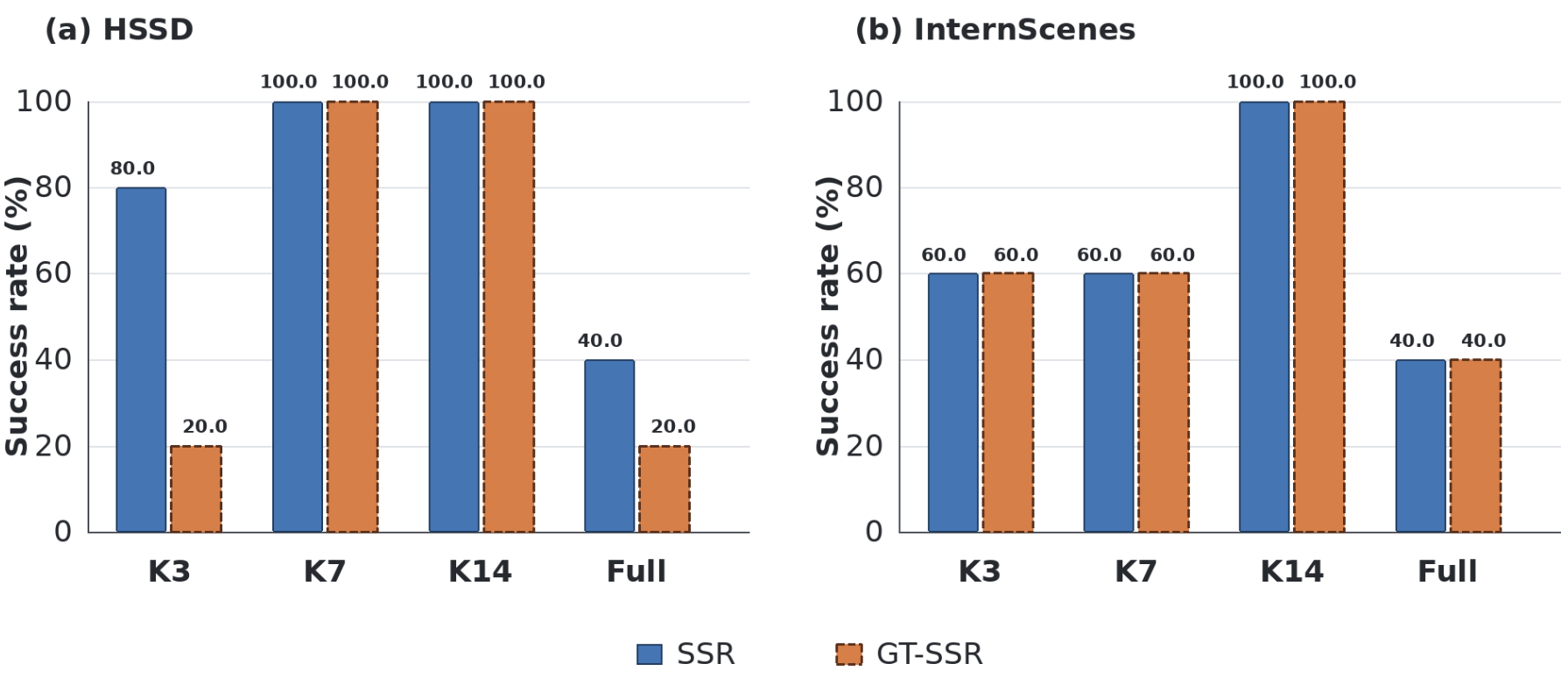}
\vspace{-2mm}
\caption{Record-capacity sensitivity of Qwen on HSSD S5 and InternScenes I3 under Estimated localization.
GT-SSR is evaluated over five matched episodes per task and capacity condition.}
\vspace{-2mm}
\label{fig:context_sensitivity}
\end{figure}

Fig.~\ref{fig:context_sensitivity} shows a nonmonotonic relationship between visible record count and GT-SSR across the two tested relation-goal tasks.
K14 is the only tested capacity achieving a GT-SSR of 1.0 on both tasks, whereas smaller prefixes can exclude useful alternatives and Full may include less relevant evidence.
This observation does not establish a universal optimal capacity or isolate the effect of individual candidate records.
Eight Qwen Full episodes terminated following context-length overflow, whereas no such failures occurred under K14.
Together with length-triggered prompt adaptation, these failures prevent attributing the observed Full--K14 differences solely to record selection or improved policy reasoning.
The corresponding reasoning cost and decision latency are examined in Sec.~\ref{sec:runtime_analysis}.

\begin{figure*}[!t]
\centering
\includegraphics[width=0.95\textwidth]{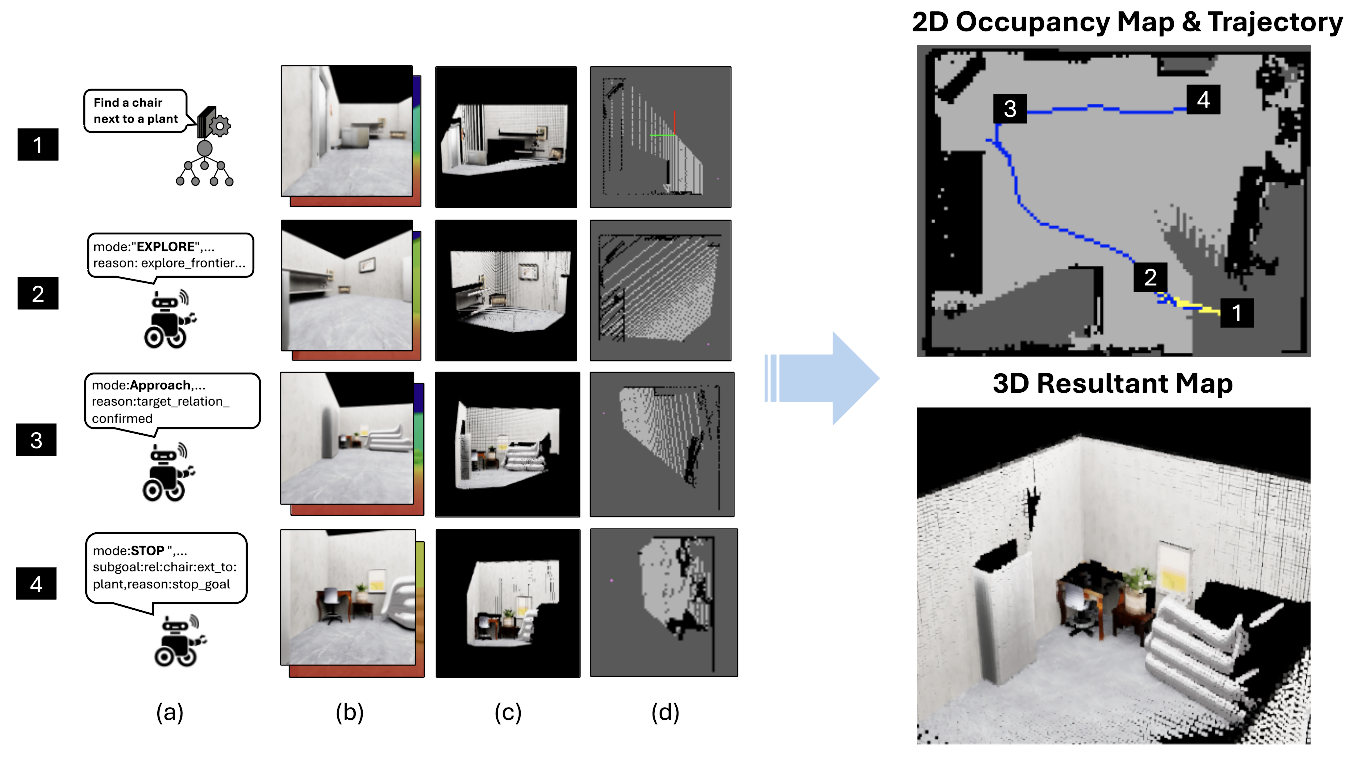}
\caption{Qualitative sequence of S\textsuperscript{3}-Nav for the relation-goal instruction ``Find a chair next to a plant.''
Four stages illustrate goal interpretation, frontier-guided exploration, target approach following relation confirmation, and terminal validation.
Panels (a)--(d) show the goal and high-level decisions, RGB-D observations, intermediate 3D reconstructions, and partial occupancy maps, respectively.
The final occupancy map shows the executed trajectory with markers 1--4, alongside the accumulated 3D reconstruction.}
\label{fig:qualitative_navigation}
\end{figure*}

\subsection{Localization and Surface-Range Diagnostics}
\label{sec:localization_diagnostics}

As a secondary diagnostic, we examine whether improved pose accuracy corresponds to better navigation outcomes and target-surface range estimates under the Qwen policy with a fixed K14 decision view. 
Table~\ref{tab:localization_comparison} shows that improved pose accuracy does not consistently translate into better navigation success or lower target-surface range error.
The surface-range diagnostics further indicate that pose estimation accuracy and target-associated geometric measurements capture different aspects of navigation reliability.
Because the measurements are obtained along condition-specific trajectories, these comparisons are descriptive and do not isolate a causal effect of localization accuracy.

\subsection{Qualitative Navigation Decision Analysis}
\label{sec:qualitative_mapping}

Fig.~\ref{fig:qualitative_navigation} illustrates how S\textsuperscript{3}-Nav uses accumulated spatial-semantic evidence to support different navigation decisions during the relation-goal task ``Find a chair next to a plant.''
While the requested chair--plant relation remains unconfirmed, the policy relies on selected frontier records for exploration as new observations are incorporated into the partial map and persistent memory. 
The chair and plant are first detected and localized as object records, with the chair serving as a potential approach anchor.
Geometric verification of the chair--plant relation then confirms that the pair satisfies the requested spatial predicate, allowing the policy to shift from exploration to target-directed approach.
Importantly, the evidence supporting this transition does not independently establish task completion, which additionally requires recent target-associated range evidence evaluated by the execution layer.

\subsection{Component Ablation}
\label{sec:component_ablation}

The ablation study examines two functions of the structured decision state, namely preserving object and relation evidence across decision steps and explicitly exposing frontier records to the high-level policy.
We use Qwen K14 under Estimated localization on the unseen HSSD relation-goal scenario with a distractor, for which the robot must combine observations acquired at different locations.
The \emph{w/o Persistent Memory} condition retains current-frame object and relation evidence but removes cross-step object/relation persistence and stale active-target hints.
The \emph{w/o Explicit Frontier Records} condition removes frontier records from the decision view but retains the partial-map and navigation summaries, including frontier availability. 
All conditions share the remaining navigation and terminal-validation procedures.

Table~\ref{tab:component_ablation} indicates that cross-step memory has a particularly important role in the evaluated unseen relation-goal task.
In one inspected execution trace from the memory ablation, the goal relation becomes confirmed while the robot is still outside the allowed terminal range.
The corresponding evidence is then unavailable at the next decision step when persistence is disabled, even though the robot has not yet completed its approach.
This failure mode distinguishes one-step relation confirmation from retaining the confirmed relation and its associated target anchor long enough to navigate to an admissible stopping position. 
This observation supports the role of persistent object--relation memory in the evaluated unseen relation-goal task.

Removing explicit frontier records yields a smaller GT-SSR decline than removing persistent memory, consistent with the continued availability of map and exploration summaries.
However, this single-task result does not establish a general ranking of component importance.
The execution layer remains unchanged across all ablation conditions, preventing this comparison from isolating its contribution to navigation performance.

\subsection{Runtime Analysis}
\label{sec:runtime_analysis}

\begin{table*}[!t]
\centering
\caption{Reasoning cost and runtime by policy, decision-view capacity, and pose-source condition across HSSD and InternScenes.
HSSD has 25 episodes per condition except for 24 valid Heuristic K14 runtime records, and InternScenes has 15 episodes per condition.
Qwen call and token counts exclude goal-schema parsing, while GT Odometry rows provide pose-source diagnostics rather than record-capacity comparisons.}
\label{tab:runtime_analysis}
\small
\setlength{\tabcolsep}{4pt}
\renewcommand{\arraystretch}{1.12}
\begin{tabular}{llccccc}
\toprule
Suite
& Condition
& \begin{tabular}[c]{@{}c@{}}Policy Lat.\\(ms) $\downarrow$\end{tabular}
& \begin{tabular}[c]{@{}c@{}}E2E Decision Lat.\\(ms) $\downarrow$\end{tabular}
& \begin{tabular}[c]{@{}c@{}}Qwen Calls/\\Ep. $\downarrow$\end{tabular}
& \begin{tabular}[c]{@{}c@{}}Tokens/Ep.\\($\times10^3$) $\downarrow$\end{tabular}
& RTF $\uparrow$ \\
\midrule

\multirow{5}{*}{HSSD}
& Heuristic Full (Est.)
& $\mathbf{0.02\pm0.01}$
& $\mathbf{30.9\pm21.8}$
& -- & --
& $0.340\pm0.041$ \\

& Heuristic K14 (Est.)
& $0.20\pm0.03$
& $33.0\pm22.5$
& -- & --
& $\mathbf{0.345\pm0.041}$ \\

& Qwen Full (Est.)
& $1535.3\pm130.0$
& $1567.4\pm138.3$
& $29.7\pm15.4$
& $86.3\pm47.2$
& $\mathbf{0.243\pm0.021}$ \\

& Qwen K14 (GT Odometry)
& $1475.3\pm110.2$
& $1594.3\pm139.4$
& $31.7\pm19.2$
& $74.5\pm47.2$
& $0.235\pm0.018$ \\

& Qwen K14 (Est.)
& $\mathbf{1486.8\pm118.2}$
& $\mathbf{1525.4\pm133.4}$
& $\mathbf{28.5\pm17.4}$
& $\mathbf{67.3\pm42.0}$
& $0.237\pm0.021$ \\

\midrule

\multirow{5}{*}{InternScenes}
& Heuristic Full (Est.)
& $\mathbf{0.022\pm0.008}$
& $29.5\pm3.0$
& -- & --
& $0.336\pm0.049$ \\

& Heuristic K14 (Est.)
& $0.214\pm0.023$
& $\mathbf{29.1\pm2.1}$
& -- & --
& $\mathbf{0.346\pm0.045}$ \\

& Qwen Full (Est.)
& $1491.7\pm103.3$
& $1537.9\pm113.2$
& $43.9\pm26.2$
& $132.9\pm87.8$
& $\mathbf{0.236\pm0.028}$ \\

& Qwen K14 (GT Odometry)
& $1483.9\pm103.0$
& $1574.3\pm106.2$
& $41.8\pm18.4$
& $95.4\pm44.3$
& $0.240\pm0.020$ \\

& Qwen K14 (Est.)
& $\mathbf{1491.3\pm115.0}$
& $\mathbf{1524.7\pm120.6}$
& $\mathbf{42.6\pm23.9}$
& $\mathbf{99.2\pm57.6}$
& $0.227\pm0.014$ \\

\bottomrule
\end{tabular}
\end{table*}
Table~\ref{tab:runtime_analysis} examines whether limiting policy-visible records reduces language-model reasoning demand and how this relates to end-to-end decision latency and simulated throughput.
Policy Latency covers the high-level policy routine, including prompt preparation, inference, and applicable response handling for Qwen.
E2E Decision Latency additionally includes perception, mapping, memory updates, and deterministic execution at policy-invoked steps, excluding subsequent simulator stepping.
The real-time factor (RTF) measures simulated episode time relative to wall-clock time.
Qwen calls and tokens exclude cached goal-schema parsing, and Tokens/Ep. includes both prompt and generated completion tokens.
The GT Odometry rows provide additional pose-source context and are excluded from the primary Full--K14 capacity comparison.

Under Estimated localization on HSSD, K14 reduces Qwen Tokens/Ep. by $22.0\%$ relative to Full, while mean Policy Latency and E2E Decision Latency decrease by only $3.2\%$ and $2.7\%$, respectively.
InternScenes shows the same qualitative separation between token consumption and per-decision latency.
These results show that K14 reduces language-model token consumption relative to Full under the implemented inference interface, but this reduction does not translate into improved simulated throughput as measured by RTF.
The heuristic has much lower inference overhead, but serves here as an alternative decision rule using the same structured evidence rather than a competing external navigation baseline.
Total token consumption is also influenced by the number of decisions along each trajectory and by length-triggered prompt adaptation, so the observations do not isolate prefix truncation as the sole source of efficiency gains.

\section{Discussion and Limitations}
\label{sec:discussion}
For the evaluated Qwen policy, K14 maintained or improved navigation success relative to Full while reducing token consumption under the implemented inference interface.
However, length-triggered prompt adaptation and context-length effects prevent attributing the observed differences solely to record capacity.
The component ablation highlights the importance of cross-step evidence retention but is limited to one unseen relation-goal scenario.
The execution layer independently validates actions and termination, although its individual contribution is not isolated because it is shared across experimental conditions.
Finally, evaluation is limited to simulation, a single Qwen model, and a small set of navigation tasks, leaving broader generalization and real-world transfer for future work.

\section{Conclusion}
\label{sec:conclusion}

This paper presented S\textsuperscript{3}-Nav, a structured spatial-semantic decision-making framework for language-guided navigation in previously unmapped environments.
The framework separates persistent object--relation memory from the capacity-controlled decision view presented to the high-level policy, preserving accumulated information for subsequent grounding and verification.
It further distinguishes provisional navigation decisions from authorized task completion by requiring confirmed goal-matching relations and recent target-associated range evidence for relational termination.
Experiments across HSSD and InternScenes showed that the K14 view maintained or improved navigation success relative to Full for the evaluated Qwen policy while reducing language-model token consumption.
The capacity analysis and memory ablation further demonstrated that exposing more records does not consistently improve navigation, whereas losing previously accumulated relational evidence can prevent successful task completion.
These findings highlight the importance of organizing spatial-semantic evidence according to its distinct roles in long-term retention, immediate decision making, and authorized task completion.

\bibliographystyle{IEEEtran}
\bibliography{refs}

\end{document}